# DOMINO++: Domain-aware Loss Regularization for Deep Learning Generalizability

Skylar E. Stolte[1], Kyle Volle[2], Aprinda Indahlastari[3,4], Alejandro Albizu[3,5], Adam J. Woods[3,4,5], Kevin Brink[6], Matthew Hale[2], and Ruogu Fang[1,3,7,8*]

[1] J. Crayton Pruitt Family Department of Biomedical Engineering, Herbert Wertheim College of Engineering, University of Florida (UF), USA
[2] Torch Technologies, LLC, Florida, USA
[3] Center for Cognitive Aging and Memory, McKnight Brain Institute, UF, USA
[4] Department of Clinical and Health Psychology, College of Public Health and Health Professions, UF, USA
[5] Department of Neuroscience, College of Medicine, UF, USA
[6] United States Air Force Research Laboratory, Eglin Air Force Base, Florida, USA
[7] Department of Electrical and Computer Engineering, Herbert Wertheim College of Engineering, UF, USA
[8] Department of Computer Information and Science and Engineering, Herbert Wertheim College of Engineering, UF, USA

**Abstract.** Out-of-distribution (OOD) generalization poses a serious challenge for modern deep learning (DL). OOD data consists of test data that is significantly different from the model's training data. DL models that perform well on in-domain test data could struggle on OOD data. Overcoming this discrepancy is essential to the reliable deployment of DL. Proper model calibration decreases the number of spurious connections that are made between model features and class outputs. Hence, calibrated DL can improve OOD generalization by only learning features that are truly indicative of the respective classes. Previous work proposed domain-aware model calibration (DOMINO) to improve DL calibration, but it lacks designs for model generalizability to OOD data. In this work, we propose DOMINO++, a dual-guidance and dynamic domain-aware loss regularization focused on OOD generalizability. DOMINO++ integrates expert-guided and data-guided knowledge in its regularization. Unlike DOMINO which imposed a fixed scaling and regularization rate, DOMINO++ designs a dynamic scaling factor and an adaptive regularization rate. Comprehensive evaluations compare DOMINO++ with DOMINO and the baseline model for head tissue segmentation from magnetic resonance images (MRIs) on OOD data. The OOD data consists of synthetic noisy and rotated datasets, as well as real data using a different MRI scanner from a separate site. DOMINO++'s superior performance demonstrates its potential to improve the trustworthy deployment of DL on real clinical data.

**Keywords:** Image Segmentation · Machine Learning Uncertainty · Model Calibration · Model Generalizability · Whole Head MRI

* Corresponding author: ruogu.fang@ufl.edu

# 1    Introduction

Large open-access medical datasets are integral to the future of deep learning (DL) in medicine because they provide much-needed training data and a method of public comparison between researchers [20]. Researchers often curate their data for DL models; yet, even the selection process itself may contain inherent biases, confounding factors, and other "hidden" issues that cause failure on real clinical data [1]. In DL, out-of-distribution (OOD) generalizability refers to a model's ability to maintain its performance on data that is independent of the model's development [23]. OOD generalizability represents a critical issue in DL research since the point of artificial intelligence (AI) in medicine is to be capable of handling new patient cases. However, this important aspect of DL is often not considered. On the other hand, overcoming challenges such as scanner-induced variance are critical in the success of neuroimaging studies involving AI [5].

We hypothesize that adaptable domain-aware model calibration that combines expert-level and data-level knowledge can effectively generalize to OOD data. DL calibration is correlated with better OOD generalizability [21]. Calibrated models may accomplish this by learning less spurious connections between features and classes. This observation relates to how calibrated models reflect the true likelihood of a data point for a class. A calibrated model may let a confusing data point naturally lay closer to the class boundaries, rather than forcing tight decision boundaries that over-fit points. Calibration affects decision-making such that the models can better detect and handle OOD data [19].

In this work, we introduce DOMINO++, an adaptable regularization framework to calibrate DL models based on expert-guided and data-guided knowledge. DOMINO++ builds on the work DOMINO[18] with three important contributions: 1) combining expert-guided and data-guided regularization to fully exert the domain-aware regularization's potential. 2) Instead of using static scaling, DOMINO++ dynamically brings the domain-aware regularization term to the same order of magnitude as the base loss across epochs. 3) DOMINO++ adopts an adaptive regularization scheme by weighing the domain-aware regularization term in a progressive fashion. The strengths of DOMINO++'s regularization lie in its ability to take advantage of the benefits of both the semantic confusability derived from domain knowledge and data distribution, as well as its adaptive balance between the data term and the regularization strength. This work shows the advantages of DOMINO++ in a segmentation task from magnetic resonance (MR) images. DOMINO++ is tested in OOD datasets including synthesized noise additions, synthesized rotations, and a different MR scanner.

## 2    Dynamic Framework for DL Regularization

### 2.1    DL backbone

U-Net transformer (UNETR) [10] serves as the DL backbone. UNETR is inspired by the awe-inspiring results of transformer modules in Natural Language Processing [22]. These modules use self-attention-based mechanisms to learn language range sequences better than traditional fully convolutional networks (FCNs). UNETR employs a transformer module as its encoder, whereas its decoder is an FCN like in the standard U-Net. This architecture learns three-dimensional (3D) volumes as sequences of one-dimensional (1D) patches. The FCN decoder receives the transformer's global information via skip connections and concatenates this information with local context that eventually recovers the original image dimensions. The baseline model does not include advanced calibration. However, basic principles to improve OOD generalizability are still incorporated for a more meaningful comparison. These principles include standard data augmentations like random Gaussian noise, rotations along each axis, and cropping. The model includes 12 attention heads and a feature size of 16.

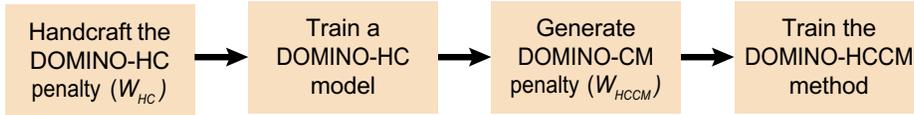

Fig. 1: The flowchart for the DOMINO++-HCCM pipeline

### 2.2  DOMINO++ Loss Regularization

**Derivation** The original DOMINO's loss regularization is as follows:

$$L(y, \hat{y}) + \beta y^T W \hat{y}, \text{ where } W = W_{HC} \text{ or } W_{CM} \tag{1}$$

where $L$ can be any uncalibrated loss function (e.g., *DiceCE* which is a hybrid of cross-entropy and Dice score [12]). $y$ and $\hat{y}$ are the true labels and model output scores, respectively. $\beta$ is an empirical static regularization rate that ranges between 0-1, and $s$ is a pre-determined fixed scaling factor to balance the data term and the regularization term. The penalty matrix $W$ has dimensions $N \times N$, where $N$ is the number of classes. $W_{HC}$ and $W_{CM}$ represent the hierarchical clustering (HC)-based and confusion matrix (CM)-based penalty matrices.

We improved its loss function to DOMINO++'s dual-guidance penalty matrix with adaptive scaling and regularization rate as follows:

$$(1 - \beta)L(y, \hat{y}) + \beta y^T (s W_{HCCM}) \hat{y} \tag{2}$$

where $\beta$ dynamically changes over epochs. $s$ is adaptively updated to balance the data and regularization terms. $W_{HCCM}$ is the dual-guidance penalty matrix.

**Combining expert-guided and data-guided regularization** DOMINO-HC regularizes classes by arranging them into hierarchical groupings based on domain. DOMINO-HC is data-independent and thus immune to noise. Yet, it becomes less useful without clear hierarchical groups. DOMINO-CM calculates class penalties using the performance of an uncalibrated model on a held-out dataset. The CM method does not require domain knowledge, but it can be more susceptible to messy data. Overall, DOMINO-HC is expert-crafted and DOMINO-CM is data-driven. These approaches have complementary advantages and both perform very well on medical image segmentation [18]. Hence, this work combines these methods to learn from experts and data.

The combined regulation (a.k.a. DOMINO-HCCM) requires first replicating DOMINO-HC. For this step, we recreate the exact hierarchical groupings from the DOMINO paper [18]. A confusion matrix is generated using DOMINO-HC on an additional validation set for matrix penalty. Next, the confusion matrix is normalized by the number of true pixels in each class. The normalized terms are subtracted from the identity matrix. Finally, all diagonals are set to 0's. Next, a second DL model trains using the resulting penalty matrix in its regularization. This process differs from DOMINO-CM because DOMINO-HC was used to generate the final model's matrix penalty. The uncalibrated model may produce a matrix penalty that is susceptible to variable quality depending on the model's training data. In comparison, the initial regularization term adds an inductive bias in the first model that encodes more desirable qualities about the class mappings [13]. Namely, the initial model contains information about the hierarchical class groupings that drives the generation of the second model's matrix penalty. The final model can now use a regularization term that is more based on task than dataset. Figure 2 displays the final DOMINO++-HCCM matrix.

**Dynamic Scaling Term** DOMINO++ adds a domain-aware regularization term to any standard loss. The resulting loss function combines the standard loss's goal of increasing accuracy with DOMINO++'s goal of re-weighing the importance of different class mix-ups when incorrect. DL models are at risk of being dominated by a specific loss during training if the losses are of different scales [14]. DOMINO [18] neglects to account for this and provides a static scaling for the regularization term based on the first epoch standard loss. In comparison, DOMINO++ updates the scaling on the regularization term to be within the same scale as the current epoch standard loss. Specifically, the scaling is computed based on the closest value to the baseline loss on the log scale. For example, an epoch with $L=13$ will have a scaling factor $S=10$.

Fig. 2: Raw matrix penalty (W) for the combined method DOMINO-HCCM. Abbreviations - BG: Background, WM: White Matter, GM: Grey Matter, CSF: Cerebrospinal Fluid, CaB: Cancellous Bone, CoB: Cortical Bone.

### 2.3   Adaptive Regularization Weighting

Multiple loss functions must be balanced properly [6]. Most studies linearly balance the separate loss terms using hyper-parameters. Hyper-parameter selection is nontrivial and can greatly alter performance. Indeed, the timing of regularization during training is critical to the final performance [8]. Hence, the current work investigates the role of regularization timing in the final model performance. Equation 2 is similar to the original DOMINO equation [18]; however, the equation is modified to include a weighting term (e.g., 1- $β$) on the standard loss function. In DOMINO, the $β$ term was simply set at a constant value of 1. As shown in Equation 3, DOMINO++ weighs the loss regularization to decay ($β$) across epochs, while the standard loss is scaled reversely with regard to the regularization term (see Equation 2).

$$β = 1 - \frac{CurrentEpoch}{TotalEpochs} \quad (3)$$

## 3   Experiments and Results

### 3.1   Dataset

**Data Source** The data in this study is from a Phase III clinical trial that tests transcranial direct current stimulation to augment cognitive training for cognitive improvement. All participants are cognitively healthy older adults between 65-89 years old. The trial was approved by the Institutional Review Boards at both study sites. Both institutions collected structural T1-weighted magnetic resonance images (T1-MRIs) from all participants. One site ("Site A") used a 3-Tesla Siemens Magnetom Prisma scanner with a 64-channel head coil and the other site ("Site B") used a 3-Tesla Siemens Magnetom Skyra scanner with a 32-channel head coil. Both locations used the following MPRAGE sequence parameters: repetition time = 1800 ms; echo time = 2.26 ms; flip angle = 8°; field of view = 256 × 256 × 256 mm; voxel size = 1 mm$^3$. The proper permissions were received for use of this dataset in this work. A total of 133 participants were included, including 123 from Site A and 10 participants from Site B.

**Reference Segmentations** The T1 MRIs are segmented into 11 different tissues, which include grey matter (GM), white matter (WM), cerebrospinal fluid (CSF), eyes, muscle, cancellous bone, cortical bone, skin, fat, major artery (blood), and air. Trained labelers performed a combination of automated segmentation and manual correction. Initially, base segmentations for WM, GM, and bone are obtained using Headreco [16], while air is generated in the Statistical Parametric Mapping toolbox [2]. Afterward, these automated outputs are manually corrected using ScanIP Simpleware™. Thresholding and morphological operations are employed to differentiate between the bone compartments. Eyes, muscle, skin, fat, and blood are manually segmented in Simpleware. Finally, CSF is generated by subtracting the ten other tissues from the whole head.

**Out-of-Domain (OOD) Testing Data** Most DL work selects a testing set by splitting a larger dataset into training and testing participants. This work also incorporates "messy" or fully independent data. Thus, three additional testing datasets are used along with the traditional testing data (Site A - Clean).

Site A Noisy - MRI noise may be approximated as Gaussian for a signal-to-noise ratio (SNR) greater than 2 [9]. Therefore, this work simulates noisy MRI images using Gaussian noise of 0 mean with a variance of 0.01.

Site A Rotated - Rotated MRI data simulates other further disturbances or irregularities (e.g., head tilting) during scanning. The rotation dataset includes random rotation of 5- to 45 degrees clockwise or counter-clockwise with respect to each 3D axis. The rotation angles are based on realistic scanner rotation [15].

Site B - Site A uses a 64-channel head coil and Site B uses a 32-channel head coil. The maximum theoretical SNR of an MRI increases with the number of channels [17]. Hence, this work seeks to test the performance of a model trained exclusively on a higher channel scanner on a lower channel testing dataset. Thus, the Site A data serves as the exclusive source of the training and validation data, and Site B serves as a unique and independent testing dataset.

### 3.2 Implementation details

This study implements UNETR using the Medical Open Network for Artificial Intelligence (MONAI-1.1.0) in Pytorch 1.10.0 [4]. The Site A data is split from 123 MRIs into 93 training / 10 validation / 10 held-out validation (matrix penalty) / 10 testing. 10 images from Site B serve as an additional testing dataset. Each DL model requires 1 GPU, 4 CPUs, and 30 GB of memory. Each model is trained for 25,000 iterations with evaluation at 500 intervals. The models are trained on $256 \times 256 \times 256$ images with batch sizes of 1 image. The optimization consists Adam optimization using stochastic gradient descent. All models segment a single head in 3-4 seconds during inference.

### 3.3 Analysis Approach

This section compares the results of 11 tissue head segmentation on each of the datasets using the baseline model, the best performing DOMINO approach, and the best performing DOMINO++ approach. The results are evaluated using Dice score [3] and Hausdorff Distance [11,7]. The Dice score represents the overlap of the model outputs with the true labels. It is better when greater and is optimally 1. Hausdorff distance represents the distance between the model outputs with the true labels. It is better when lesser and is optimally 0.

### 3.4 Segmentation Results

**Qualitative Comparisons** Figure 3 features segmentation of one example slice from the Site A MRI dataset with Gaussian Noise. DOMINO substantially exaggerates the blood regions in this slice. In addition, DOMINO entirely misses a section of white matter near the eyes. However, DOMINO can also capture certain regions of the white matter, particularly in the back of the head, better than the baseline model. In general, all outputs have noisy regions where there appear to be "specks" of an erroneous tissue. For instance, grey matter is incorrectly identified as specks within the white matter. This issue is far more common in the DOMINO output compared to the baseline or DOMINO++ outputs.

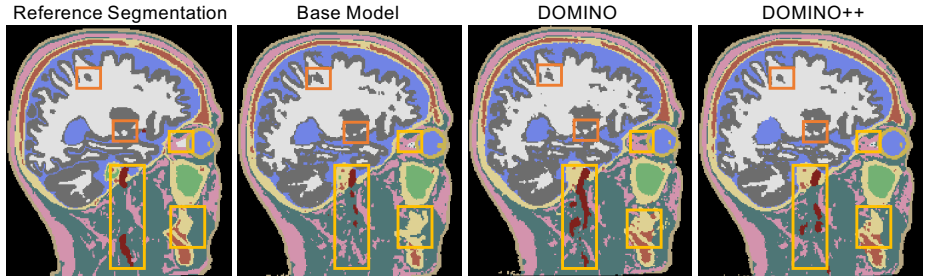

Fig. 3: Visual comparison of segmentation performance on a noisy MRI image from Site A. The yellow rectangular regions show areas where DOMINO++ improves the segmentation. The orange regions show areas that DOMINO and DOMION++ improve the segmentation over the baseline model.

**Quantitative Comparisons** Tables 1 and 2 show that DOMINO++ achieves the best Dice scores and Hausdorff Distances across all test sets, respectively. As such, DOMINO++ produces the most accurate overall segmentation across tissue types. The supplementary material provides individual results across every dataset and tissue type. So far, DOMINO++ improves the model generalizability to the noisy and rotated datasets the most. These improvements are important in combating realistic MR issues such as motion artifacts. Future work will build off of DOMINO++'s improvements on different scanner data to yield even better results. Table 3 displays the Hausdorff Distances for every tissue across Site B's test data. Site B is highlighted since that this is real-world OOD data. DOMINO++ performs better in most tissues and the overall segmentation. GM, cortical bone, and blood show the most significant differences with DOMINO++. This is highly relevant to T1 MRI segmentation. Bone is difficult to differentiate from CSF with only T1 scans due to similar contrast. Available automated segmentation tools use young adult heads as reference, whereas the bone structure between older and younger adults is very different (e.g., more porous in older adults). Hence, DOMINO++ is an important step in developing automated segmentation tools that are better suited for older adult heads.

**Ablation Testing** The supplementary material provides the results of ablation testing on DOMINO++. These results compare how $W_{HCCM}$, $s$, and $\beta$ individually contribute to the results. Interestingly, different individual terms cause the model to perform stronger in specific datasets. Yet, the combined DOMINO++ still performs the best across the majority of datasets and metrics. These ob-servations suggest that each term has strengths on different data types that can strengthen the overall performance.

Table 1: Average Dice Scores. The data is written as mean±standard deviation. * Denotes Significance using multiple comparisons tests.

| Method | Site A clean | Site A noisy | Site A rotated | Site B |
|---|---|---|---|---|
| Base | 0.808±0.014 | 0.781±0.015 | 0.727±0.041 | 0.730±0.028 |
| DOMINO | 0.826±0.014 | 0.791±0.018 | 0.777±0.023 | 0.750±0.026 |
| DOMINO++ | 0.842±0.012* | 0.812±0.016* | 0.789±0.23* | 0.765±0.027 |

Table 2: Average Hausdorff Distances. The data is written as mean±standard deviation. * Denotes Significance using multiple comparisons tests.

| Method | Site A clean | Site A noisy | Site A rotated | Site B |
|---|---|---|---|---|
| Base | 0.651±0.116 | 0.669±0.085 | 2.266±1.373 | 1.699±0.414 |
| DOMINO | 0.525±0.090 | 0.565±0.149 | 1.284±0.500 | 1.782±0.669 |
| DOMINO++ | 0.461±0.077* | 0.457±0.076* | 1.185±0.411* | 1.228±0.414 |

Table 3: Hausdorff Distances on Site B data. The data is written as mean±standard deviation. * Denotes Significance using multiple comparisons tests. Abbreviations - CaB: Cancellous Bone. CoB: Cortical Bone.

| Tissue | WM | GM | Eyes | CSF | Air | Blood | CaB | CoB | Skin | Fat | Muscle |
|---|---|---|---|---|---|---|---|---|---|---|---|
| Base | 0.215±0.063 | 0.266±0.107 | 1.089±0.881 | 0.506±0.226 | 2.281±1.426 | 8.534±2.564 | 1.581±0.626 | 2.203±1.279 | 0.610±0.424 | **0.958±0.751** | 0.363±0.159 |
| DOMINO | 0.260±0.117 | 0.221±0.048 | 1.600±4.108 | 0.564±0.198 | 2.070±1.380 | 9.934±4.025 | **1.456±0.672** | 1.331±0.827 | 0.811±0.838 | 1.040±0.987 | 0.320±0.234 |
| DOMINO++ | **0.189±0.042** | **0.171±0.036*** | **0.149±0.047** | **0.462±0.106** | **1.446±1.057** | **6.260±2.195*** | 1.996±0.960 | **0.950±0.705*** | **0.570±0.369** | 1.060±0.935 | **0.308±0.165** |

**Training Time Analysis** DOMINO-HC took about 12 hours to train whereas DOMINO-CM and DOMINO++ took about 24 hours to train. All models took 3-4 seconds per MR volume at the inference time. A task that has very clear hierarchical groups may still favor DOMINO-HC for the convenient training time. This might include a task with well-documented taxonomic levels (e.g., animal classification). However, medical data is often not as clear, which is why models that can learn from the data are valuable. DOMINO++ makes up for the longer training time by learning more specific class similarities from the data. Tasks that benefit from DOMINO++ over DOMINO-HC are those that only have loosely-defined categories. Tissue segmentation falls under this domain because tissues largely occur in similar anatomical locations (strength of DOMINO-HC) but the overall process is still variable with individual heads (strength of DOMINO-CM).

## 4  Conclusions

DOMINO [18] established a framework for calibrating DL models using the semantic confusability and hierarchical similarity between classes. In this work, we proposed the DOMINO++ model which builds upon DOMINO's framework with important novel contributions: the integration of data-guided and expert-guided knowledge, better adaptability, and dynamic learning. DOMINO++ surpasses the equivalent uncalibrated DL model and DOMINO in 11-tissue segmentation on both standard and OOD datasets. OOD data is unavoidable and remains a pivotal challenge for the use of artificial intelligence in clinics, where there is great variability between different treatment sites and patient populations. Overall, this work indicates that DOMINO++ has great potential to improve the trustworthiness and reliability of DL models in real-life clinical data. We will release DOMINO++ code to the community to support open science research.

**Acknowledgements** This work was supported by the National Institutes of Health / National Institute on Aging, USA (NIA RF1AG071469, NIA R01AG054077), the National Science Foundation, USA (1908299), the Air Force Research Laboratory Munitions Directorate, USA (FA8651-08-D-0108 TO48), and NSF-AFRL INTERN Supplement to NSF IIS-1908299, USA.

# Supplementary Material

## 1   Additional Visual Results

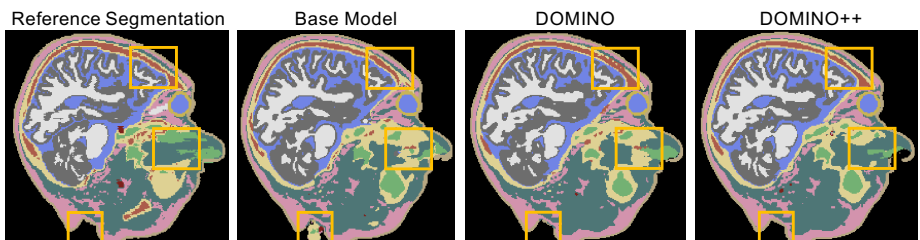

**Fig. 1.** Segmentation results on a Site A rotated testing image. The yellow squares highlight areas where DOMINO++improves the performance. DOMINO-CM is used for DOMINO in all comparisons because it performs better than DOMINO-HC.

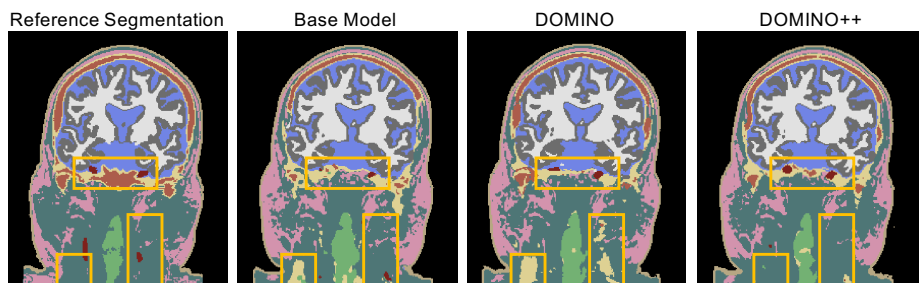

**Fig. 2.** Segmentation results on a Site B testing image. The yellow squares highlight areas where DOMINO++ improves the performance.

## 2   Performance on all Datasets

**Table 1.** Performance comparison across tissues using Dice Scores on Site A clean data. Abbreviations - CaB: Cancellous Bone. CoB: Cortical Bone.

| Tissue | WM | GM | Eyes | CSF | Air | Blood | CaB | CoB | Skin | Fat | Muscle |
|---|---|---|---|---|---|---|---|---|---|---|---|
| Base | 0.9301 | 0.8899 | 0.8253 | 0.8424 | 0.8249 | 0.3751 | 0.7113 | 0.7679 | 0.8737 | 0.9325 | 0.9104 |
| DOMINO | 0.9429 | 0.9021 | 0.8402 | 0.8424 | **0.8472** | 0.4634 | 0.7370 | 0.7962 | 0.8742 | 0.9291 | 0.9152 |
| DOMINO++ | **0.9453** | **0.9046** | **0.8623** | **0.8558** | 0.8457 | **0.5363** | **0.7607** | **0.8028** | **0.8926** | **0.9363** | **0.9202** |

**Table 2.** Performance comparison across tissues using Dice Scores on Site A with Gaussian Noise. Abbreviations - CaB: Cancellous Bone. CoB: Cortical Bone.

| Tissue | WM | GM | Eyes | CSF | Air | Blood | CaB | CoB | Skin | Fat | Muscle |
|---|---|---|---|---|---|---|---|---|---|---|---|
| Base | 0.8779 | 0.8279 | 0.8249 | 0.8069 | 0.8108 | 0.3731 | 0.6825 | 0.7533 | 0.8497 | 0.8992 | 0.8851 |
| DOMINO | 0.8872 | 0.8376 | 0.8237 | 0.8075 | **0.8345** | 0.4331 | 0.6845 | 0.7693 | 0.8387 | 0.8958 | 0.8830 |
| DOMINO++ | **0.9062** | **0.8509** | **0.8545** | **0.8212** | 0.8323 | **0.5164** | **0.7104** | **0.7805** | **0.8584** | **0.9021** | **0.8949** |

**Table 3.** Performance comparison across tissues using Dice Scores on Site A with Rotations. Abbreviations - CaB: Cancellous Bone. CoB: Cortical Bone.

| Tissue | WM | GM | Eyes | CSF | Air | Blood | CaB | CoB | Skin | Fat | Muscle |
|---|---|---|---|---|---|---|---|---|---|---|---|
| Base | 0.9094 | 0.8610 | 0.6821 | 0.7900 | 0.7357 | 0.1302 | 0.6102 | 0.6839 | 0.8226 | 0.9112 | 0.8579 |
| DOMINO | 0.9290 | 0.8825 | 0.8033 | 0.8110 | 0.7952 | 0.2721 | 0.6598 | 0.7554 | 0.8372 | 0.9148 | 0.8858 |
| DOMINO++ | **0.9306** | **0.8848** | **0.8181** | **0.8217** | **0.7986** | **0.3296** | **0.6857** | **0.7555** | **0.8464** | **0.9187** | **0.8909** |

## 3 Ablation Testing

**Table 4.** Performance comparison using average Dice Score across sites. DOMINO-CM is used as the base matrix penalty for DOMINO++ w/o [HCCM] instead of DOMINO-HC due to higher quantitative performance. S: adaptive scaling term. R: dynamic regularization weighting

| Method | Site A clean | Site A noisy | Site A rotated | Site B |
|---|---|---|---|---|
| DOMINO++ | **0.8421** | **0.8116** | **0.7891** | **0.7653** |
| DOMINO++ w/o [HCCM] | 0.8348 | 0.7980 | 0.7887 | 0.7563 |
| DOMINO++ w/o [R] | 0.8330 | 0.8069 | 0.7873 | 0.7593 |
| DOMINO++ w/o [S] | 0.8367 | 0.7944 | 0.7828 | 0.7568 |

**Table 5.** Performance comparison using average Hausdorff Distance across sites. DOMINO-CM is used as the base matrix penalty for DOMINO++ w/o [HCCM] instead of DOMINO-HC due to higher quantitative performance. S: adaptive scaling term. B: dynamic regularization weighting.

| Method | Site A clean | Site A noisy | Site A rotated | Site B |
|---|---|---|---|---|
| DOMINO++ | **0.4609** | **0.4565** | 1.1853 | **1.2279** |
| DOMINO++ w/o [HCCM] | 0.5265 | 0.5402 | **1.1497** | 1.7052 |
| DOMINO++ w/o [R] | 0.5332 | 0.5149 | 1.2195 | 1.4990 |
| DOMINO++ w/o [S] | 0.4902 | 0.5212 | 1.2317 | 1.5640 |